\def\year{2022}\relax
\documentclass[letterpaper]{article} 
\usepackage{aaai22}  
\usepackage{times}  
\usepackage{helvet}  
\usepackage{courier}  
\usepackage[hyphens]{url}  
\usepackage{graphicx} 
\usepackage{natbib}  
\usepackage{caption} 
\DeclareCaptionStyle{ruled}{labelfont=normalfont,labelsep=colon,strut=off} 
\usepackage{algorithm}
\usepackage{algorithmic}
\usepackage{amsmath}
\usepackage{amsfonts}
\usepackage{comment}
\usepackage[section]{placeins}

\graphicspath{ {./images/} }

\usepackage{todonotes}
\usepackage{soul} 
\usepackage{ulem} 
\usepackage{cancel}
\usepackage{setspace} 

\usepackage{color}
\definecolor{green}{RGB}{0,121,52}

\newcommand{\VR}[1]{{#1}}

\allowdisplaybreaks

\usepackage{newfloat}
\usepackage{listings}
\floatstyle{ruled}
\newfloat{listing}{tb}{lst}{}
\floatname{listing}{Listing}
\title{Stable Anderson Acceleration for Deep Learning}
\author{
    Massimiliano Lupo Pasini\textsuperscript{\rm 1}\equalcontrib, 
    Junqi Yin\textsuperscript{\rm 2}\equalcontrib, 
    Viktor Reshniak\textsuperscript{\rm 3}\equalcontrib, 
    Miroslav Stoyanov\textsuperscript{\rm 3}\equalcontrib
}
\affiliations{
    \textsuperscript{\rm 1}Oak Ridge National Laboratory, Computational Sciences and Engineering Division, Oak Ridge, TN, USA, 37831\\
    \textsuperscript{\rm 2}Oak Ridge National Laboratory, National Center for Computational Sciences Division, Oak Ridge, TN, USA, 37831\\
    \textsuperscript{\rm 3}Oak Ridge National Laboratory, Computer Science and Mathematics Division, Oak Ridge, TN, USA, 37831\\
    \{lupopasinim, yinj, reshniakv, stoyanovmk\}@ornl.gov
    
}

\begin{document}

\maketitle

\begin{abstract}
 Anderson acceleration (AA) is an extrapolation technique designed to speed-up fixed-point iterations like those 
arising from the iterative training of DL models. Training DL models requires large datasets processed in randomly sampled batches that tend to introduce in the fixed-point iteration stochastic oscillations of amplitude roughly inversely proportional to the size of the batch. These oscillations reduce and occasionally eliminate the positive effect of AA. 
To restore AA's advantage, we combine it with an adaptive moving average procedure that smoothes the oscillations and results in a more regular sequence of gradient descent updates. 
By monitoring the relative standard deviation between consecutive iterations, we also introduce a criterion to automatically assess whether the moving average is needed.
We applied the method to the following DL instantiations: (i) multi-layer perceptrons (MLPs) trained on the open-source \texttt{graduate admissions} dataset for regression, (ii) physics informed neural networks (PINNs) trained on source data to solve 2d and 100d Burgers' partial differential equations (PDEs), and (iii) ResNet50 trained on the open-source \texttt{ImageNet1k} dataset for image classification. Numerical results obtained using up to 1,536 NVIDIA V100 GPUs on the OLCF supercomputer Summit showed the stabilizing effect of the moving average on AA for all the problems above. 
\end{abstract}

{\footnotesize \noindent This manuscript has been authored in part by UT-Battelle, LLC, under contract DE-AC05-00OR22725 with the US Department of Energy (DOE). The US government retains and the publisher, by accepting the article for publication, acknowledges that the US government retains a nonexclusive, paid-up, irrevocable, worldwide license to publish or reproduce the published form of this manuscript, or allow others to do so, for US government purposes. DOE will provide public access to these results of federally sponsored research in accordance with the DOE Public Access Plan (\url{http://energy.gov/downloads/doe-public-access-plan}).




\section{Introduction}

Training deep learning (DL) models is computationally intensive because of the high dimensional and nonlinear nature of the problem. 
First-order optimization methods are standard training techniques due to their
low memory requirements and cost per iteration. However, the convergence rate of these techniques can be relatively slow.  
Accelerating the training of DL models is possible by exploiting information about the curvature of the loss function landscape with quasi-Newton methods that compute a low rank approximation of the Hessian. 
Quasi-Newton methods \cite{NoceWrig06} are categorized into (i) those that require additional function evaluations to approximate the Hessian, and (ii) those that use only existing information already available in the standard fixed-point method. 
\VR{The first group contains} momentum acceleration methods \cite{qian_momentum_1999, barakat_stochastic_2020} and stochastic Newton methods \cite{haber_never_2018, tong_asynchronous_2021, berahas_quasi-newton_2020, tong_asynchronous_2021}, whereas \VR{group (ii) contains} natural gradient descent (NGD) \cite{NGD, 8752867} and extrapolation techniques for vector sequences \cite{brezinski_shanks_2018, saad2021}. 
Methods in (i) generally result in faster acceleration; however, methods in (ii) are preferred when evaluating the loss function is expensive since the additional computational cost may eliminate the benefits of the quasi-Neuton apporach.
Presently, evaluating the training loss function is becoming increasingly expensive, since the nonlinear DL models are increasing in dimensionality and training has to be performed on distributed high-performance computing (HPC) platforms.
Therefore, methods in group (ii) are more suitable for accelerating the training of large DL models at scale. 

Recent studies have revealed that NGD is unlikely to accelerate the training of DL models because the empirical Fisher information matrix does not 
accurately capture second-order information \cite{kunstner2020limitations}. 
Alternatively, extrapolation techniques such as Anderson acceleration (AA)
\cite{anderson_iterative_1965, fang_two_2009, walker_anderson_2011, potra_characterization_2013, brezinski_shanks_2018, wang_asymptotic_2020, toth_local_2017} have also
gained interest in the machine learning (ML) community \cite{shi_regularized_2019, 9308288}. 
One of the known limitations of AA is that its performance is hindered by the stochastic oscillations introduced by batch processing of the training data \cite{toth_local_2017}.
To improve the efficacy of AA in presence of large stochastic oscillations, regularization is needed. 
The authors in \cite{scieur_online_2019} proposed a regularized nonlinear acceleration (RNA) that alters the numerical scheme used to perform AA, whereas other approaches \cite{mai_anderson_2020} change the training loss to enforce regularization. 
However, these approaches have been shown successful only on low dimensional logistic regression problems. 

{\it Here, we propose a stable approach to perform AA that accelerates the training of DL models}. In contrast to previous works, we regularize AA by averaging the gradients computed on consecutive first-order optimization updates \cite{polyak1991, 5694086, flammarion2014, flammarion2015, Tripuraneni2018, Sportisse2020}, \VR{which reduces the amplitude of the stochastic oscillations}. 
In particular, we use a variant of AA called Alternating Anderson \cite{banerjee_periodic_2016, pratapa_anderson_2016, suryanarayana_alternating_2019,  lupo_pasini_convergence_2019} that relaxes the frequency of AA corrections by interleaving an AA update with multiple steps of the first-order optimizer. 
Differently from AA schemes commonly used by the ML community, 
Alternating Anderson is more effective because the relaxed frequency of AA reduces the risk of co-linearity (i.e., linear dependence) between consecutive updates, which in turns helps the quasi-Newton methods to 
\VR{avoid stagnation} and escape from bad local minima. 
Our implementation of AA is open-source and publicly available at the GitHub repository \texttt{https://github.com/ORNL/AADL}.

\VR{We illustrate numerically the efficacy of the proposed acceleration technique on several problems including (i) multi-layer perceptrons (MLP) trained on the open-source \texttt{graduate admissions} dataset \cite{graduate}, (ii) physics informed neural networks (PINNs) \cite{PINNs2019} to solve partial differential equations (PDEs), and (iii) ReNet50 trained on the \texttt{ImageNet1k} \cite{imagenet} benchmark image dataset.}
The training of ResNet50 on \texttt{ImageNet1k} was performed with distributed data parallelism to cope with the larger volume of data. 
The results show that the combined use of moving average and AA succeeds in efficiently accelerating the training of DL models also in situations where AA by itself is not effective. Moreover, the results for ResNet50 trained on the \texttt{ImageNet1k} dataset with distributed data parallelism show that AA with moving average helps the DL model recover from the generalization gap \cite{generalization}.

\section{Mathematical background}
We denote with $D \in \mathbb{N}$ the total number of data points that compose the entire training dataset.
We represent the DL model as a function $F: \mathbb{R}^{c}\times \mathbb{R}^n \rightarrow \mathbb{R}^s$ that depends on two sets of parameters: the input data $\mathbf{x} \in \mathbb{R}^c$ and the weights $\mathbf{w}\in \mathbb{R}^n$. The prediction produced by the DL model  for a given input data point $\mathbf{x}$ and weights $\mathbf{w}$ is $F(\mathbf{x},\mathbf{w}) \in \mathbb{R}^s$.
  
The loss function $L:\mathbb{R}^n\rightarrow \mathbb{R}$ measures the mismatch between output data $\mathbf{y}$ and the values $F(\mathbf{x},\mathbf{w})$ predicted by the DL model, 
and we aim at computing $\mathbf{w}_{\text{opt}}\in \mathbb{R}^n$ such that $L(\mathbf{w})$ is minimized. Since the DL model is nonlinear, computing $\mathbf{w}_{\text{opt}}$ is performed in an iterative fashion via gradient descent optimization techniques. Computing a gradient descent update on the entire dataset is not computationally convenient due to the typically large volume of the training data; therefore, the dataset is split into batches that are processed separately, and each batch is employed to compute a new correction on the numerical approximation to $\mathbf{w}_{\text{opt}}$.
We define $\mathcal{B}_k \subset \{1,\ldots,D\}$ as the set of indices that identify the data points belonging to the specific data batch processed at step $k$, and $B\le D$ denotes the cardinality of $\mathcal{B}_k$ for every $k$. The binary vector
$\boldsymbol{\xi}^k \in \mathbb{R}^D$ is defined such that $\xi_i = 1$ if $i\in B$ and $\xi_i = 0$ if $i\not\in B$, and is used to randomly select the batch at step $k$.    

We define $\ell(\cdot, \boldsymbol{\xi}^k)$ as the restriction of the loss function $L(\cdot)$ on the data batch identified by $\mathcal{B}_k$. The gradient $\nabla_{\mathbf{w}}\ell(\mathbf{w}^k, \boldsymbol{\xi}^k)$ is used to update the weights $\mathbf{w}^k$. 
The stochastic optimization scheme that subsequently minimizes
$\ell(\mathbf{w}^k,\boldsymbol{\xi}^k)$ on different data batches identified by $\mathcal{B}_k$, can be formulated as a stochastic nonlinear fixed-point iteration
\begin{equation}
\begin{split}
\mathbf{w}^{k+1} & = G_k(\mathbf{w}^{k}, \boldsymbol{\xi}^k)\\& = \mathbf{w}^k + \mathbf{r}^k, \quad k=0,1,\ldots
\end{split}
\label{fixed_point}
\end{equation}
where $G_k:\mathbb{R}^n \rightarrow \mathbb{R}^n$ is the stochastic nonlinear fixed-point operator, e.g., the forward-backward pass of a stochastic first-order optimizer to update the parameters of a DL model. The vectors $\mathbf{w}^{k}\in \mathbb{R}^{n}$ are consecutive updates of the weights of the DL model, and $\mathbf{r}^k$ is residual of the fixed-point iteration, which depends on the specific stochastic optimizer. 
The definition of the fixed-point operator $G_k$ depends on the data batch identified by $\mathcal{B}_k$; therefore, the fixed-point operator $G_k$ generally varies across the iterative training of the DL model, which makes the fixed-point scheme described in Equation \eqref{fixed_point} non-stationary \cite{Nonlinear_fixed_point}. 

\subsection{Anderson acceleration}
The original AA method was proposed in \cite{anderson_iterative_1965}.
Define the rectangular matrices
\begin{equation}
\begin{split}
W_k = & [(\mathbf{w}^{k-m+1}-\mathbf{w}^{k-m}), (\mathbf{w}^{k-m+2}-\mathbf{w}^{k-m+1}), \ldots, \\ & (\mathbf{w}^{k}-\mathbf{w}^{k-1})]\in \mathbb{R}^{n\times m}
\end{split}
\label{stacking_x}
\end{equation}
and 
\begin{equation}
\begin{split}
R_k = & [(\mathbf{r}^{k-m+1}-\mathbf{r}^{k-m}), (\mathbf{r}^{k-m+2}-\mathbf{r}^{k-m+1}), \ldots, \\& (\mathbf{r}^{k}-\mathbf{r}^{k-1})]\in \mathbb{R}^{n\times m}.
\end{split}
\label{stacking_r}
\end{equation}
The dimension $m$ is a tunable parameter that determines the number of past iterates to extrapolate a new converging sequence.
The goal is to compute a vector $\mathbf{g}^k\in \mathbb{R}^m$ to 
correct the approximation $\mathbf{w}^k$ as follows:
\begin{equation}
\overline{\mathbf{w}}^k = \mathbf{w}^k + \mathbf{r}^k - (W_k+R_k)\mathbf{g}^k,
\label{anderson_acceleration}
\end{equation}
where the optimal vector $\mathbf{g}^k$ is computed as
\begin{equation}
\mathbf{g}^k = \underset{\mathbf{g}\in \mathbb{R}^{m}}{\operatorname{argmin}}\lVert 
\mathbf{r}^k-R_k\mathbf{g}\rVert^2_{2}.
\label{least_squares}
\end{equation}
The procedure proposed in Equation \eqref{anderson_acceleration} to 
compute $\overline{\mathbf{w}}^{k}$ is called AA, and it can be interpreted as a step of a \textit{projection method} (see Chapter 5 of \cite{Saad2003}). 
We then perform a linear mixing between the last computed weight update $\mathbf{w}^k$ and the Anderson accelerated $\overline{\mathbf{w}}^k$ using a scalar relaxation parameter $\beta$ as follows
\begin{equation}
    \mathbf{w}^{k+1} = (1-\beta)\mathbf{w}^k + \beta \overline{\mathbf{w}}^k, \qquad 0\le \beta \le 1.
\end{equation}
AA may become numerically unstable if it is performed at each iteration, because the columns of the matrix $R_k$ tend to be co-linear and the least-squares problem in Equation \ref{least_squares} becomes ill-conditioned. 

To reduce the risk of co-linearity in the columns of $R_k$, we consider in this work a variant of AA called \textit{Alternating Anderson}, which was originally proposed in \cite{pratapa_anderson_2016} and whose convergence on linear problems is analyzed in \cite{lupo_pasini_convergence_2019}. Here we generalize this algorithm to nonlinear high dimensional stochastic optimization problems like the ones arising from the training of DL models. The idea of Alternating Anderson is to relax the frequency with which AA is performed waiting for multiple stochastic gradient updates to be computed, so that the cost of solving successive least-squares problems is amortized by several inexpensive stochastic gradient updates in between. 
We call this approach \textit{Anderson Acceleration for Deep Learning (AADL)}.

The iteration of AADL has the following form 
\begin{equation}
\mathbf{w}^{k+1} = \mathbf{w}^k + C_k \mathbf{r}^{k},
\label{alternating_anderson_ck}
\end{equation}
where the $n\times n$ matrix $C_k$ is defined as
\begin{equation}
C_k = \begin{cases}
I, \qquad \qquad \qquad \qquad \qquad \qquad \qquad \,\;\; k/p \notin \mathbb{N} \\
(1-\beta) I - \beta (W_k + R_k)(R_k^T R_k)^{-1} R_k^T,  k/p \in \mathbb{N}.
\end{cases}
\end{equation}
The tunable parameter $p$ represents the number of stochastic gradient updates separating two consecutive Anderson accelerations.
Letting $p\rightarrow \infty$ would reduce the numerical scheme to the original stochastic optimization technique, while $p=1$ would perform AA at each optimization step. Similarly to relaxing the frequency to perform AA, one could also relax the frequency to update the tall and skinny matrices $W_k$ and $R_k$, and we denote this storage frequency with $q$. 
The value of $q$ does not need to be related to $p$, but we recommend that $q\le p$. The iterative scheme in Equation \eqref{alternating_anderson_ck} is repeated until the loss function evaluated on a data batch drops below a tolerance $tol$ defined by the user.

\subsection{Moving average regularization}
For convenience in describing the mathematical formulation of our approach, we consider only batch gradient updates of the form 
\begin{equation}
\mathbf{r}^{k-1} = \alpha\nabla \ell (\mathbf{w}^{k-1}, \boldsymbol{\xi}^{k-1}),
\end{equation}
 where the scalar positive tuneable parameter $\alpha$ is the learning rate of the stochastic optimizer.
The update $\mathbf{r}^k$ is an unbiased statistical estimator of $\nabla L(\mathbf{x}^k)$, meaning that the expected value $\mathbb{E}[\mathbf{r}^k] = \nabla L(\mathbf{x}^k)$. 
Moreover, we assume that the stochastic error $\boldsymbol{\eta}^{k}(\mathbf{w}^k)\in \mathbb{R}^{n}$ is an additive noise with zero-mean and bounded noise variance: 
\begin{equation}
    \nabla \ell(\mathbf{w}^k,\boldsymbol{\xi}^{k})= \nabla L (\mathbf{w}^{k})+\boldsymbol{\eta}^{k}(\mathbf{w}^k).
    \label{additive_noise}
\end{equation}
We also assume that the values of the random noise $\boldsymbol{\eta}^k$ are uncorrelated between different gradient update steps.
In general, early iterations of the stochastic optimization are characterized by large values of the variance, and this prevents any AA scheme as the one described in Algorithm \ref{AAR_algorithm} from effectively accelerating the convergence of stochastic fixed-point iterations \cite{toth_local_2017}. Therefore, a regularization technique is needed to reduce the variance of the stochastic gradient updates and transform the original highly oscillatory sequence of gradient updates into a smoother one, on which AA can efficiently be used as an acceleration.

To this effect, we extend the moving average method \cite{polyak1991, 5694086, flammarion2014, flammarion2015, Tripuraneni2018, Sportisse2020} originally used on first-order optimizers to regularize the sequence $\{\mathbf{w}^k\}_{k=0}^{\infty}$ generated by a quasi-Newton stochastic optimization algorithm like AADL.
This approach generates a new averaged sequence $\{ \tilde{\mathbf{w}}^k\}_{k=1}^{\infty}$ that stems from the original one according to the following formula
\begin{equation}
\begin{split}
   \tilde{\mathbf{w}}^{k} = \frac{1}{t}\sum_{i=0}^{t-1}  \mathbf{w}^{k-i}
   = \frac{1}{t}\sum_{i=k-t+1}^{k}  \mathbf{w}^{i}.
   \label{moving_average}
\end{split}
\end{equation}
where we pick $t<p$, meaning that the average window does not exceed the sliding window of successive gradient updates used to perform AA.  
Using Equation \eqref{additive_noise} we obtain
\begin{equation}
\begin{split}
    \tilde{\mathbf{w}}^{k} = \mathbf{w}^k &+ \underbrace{\sum_{i=1}^{t-1} \frac{t-i}{t} \nabla L(\mathbf{w}^{k-i})}_\text{smoothing}  + \underbrace{\sum_{i=1}^{t-1}
    \frac{t-i}{t} \boldsymbol{\eta}^{k-i}(\mathbf{w}^{k-i})}_\text{averaging of the noise}
    \label{benefits}
\end{split}
\end{equation}
Equation \eqref{benefits} highlights the dual benefit of the moving average. 
The first sum on the right hand side of Equation \eqref{benefits}
represents a smoothing over successive gradient updates. Since the loss function $L$ can have very steep local minima, which can cause the DL model to overfit, the smoothing obtained by the moving average can help the training of the DL model to escape from bad local minima and continue exploring the loss function landscape.
The second sum on the right hand side
represents an average over the additive stochastic noise. Using the assumption that the values of the random noise $\boldsymbol{\eta}^k$ at different steps $k$ are uncorrelated, we obtain
\begin{equation}
\begin{split}
\text{Var}(\tilde{\mathbf{w}}^k) = \sum_{i=1}^{t-1} \frac{(t-i)^2}{t^2} \text{Var}\bigg [\boldsymbol{\eta}^{k-i}(\mathbf{w}^{k-i})\bigg ].
\end{split}
\label{variance_tilde}
\end{equation}
Since the variance of the original sequence of weights $\mathbf{w}^k$ (under the same assumptions) is
\begin{equation}
\begin{split}
    \text{Var}(\mathbf{w}^k) = \sum_{i=1}^t \text{Var}\bigg [\boldsymbol{\eta}^{k-i}(\mathbf{w}^{k-i})\bigg ]
\label{variance}
\end{split}
\end{equation}
a comparison between Equation \eqref{variance_tilde} and Equation \eqref{variance} shows that $\text{Var}(\tilde{\mathbf{w}}^k) < \text{Var}(\mathbf{w}^k)$. Therefore,
the averaging of the noise performed by the moving average approach benefits the training by reducing the variance of the stochastic gradient updates, which in turn dampens the stochastic oscillations and favors the use of AA. 
Although increasing the value of $t$ leads to a further reduction of the variance, using an excessively large value for $t$ would average over too many updates and thus deteriorate the convergence rate of the training.
A compromise between variance reduction and convergence rate has to be found to achieve the best convergence acceleration.
To this effect, we propose a criterion to automatically select whether to perform the moving average or not based on the standard deviation of the entries of the vector $\mathbf{w}^k$ across successive gradient descent updates. 

The sample variance $S(\mathbf{w}^k) \in \mathbb{R}^n$ is defined entry-wise as follows:
\begin{equation}
\begin{split}
    S_j(\mathbf{w}^k) & = \frac{1}{t} \sum_{i=0}^{t-1} (\mathbf{w}^{k-i})^2_j
    - \bigg [ \frac{1}{t} \sum_{i=0}^{t-1} (\mathbf{w}^{k-i})_j \bigg ]^2, \\& \quad j=1,\ldots,n
\end{split}
\end{equation}
and we use it in our adaptive criterion as follows:
\begin{equation}
    \left\|\sqrt{S(\mathbf{w}^k)}\right\|_{\infty} < \varepsilon \cdot \|\mathbf{r}^k\|_{\infty}
    \label{stopping_criterion}
\end{equation}
where the square root operation $\sqrt{\cdot}$ is performed entry-wise, $\|\cdot\|_{\infty}:=\max_{j=1, \ldots, n}|\cdot|$, and $\varepsilon > 0$ is a tuneable scalar parameter. We stop performing the moving average whenever the weights in the vector $\mathbf{w}^k$ satisfy the condition in Equation \eqref{stopping_criterion}. 
 Larger values of $\varepsilon$ reduce the frequency of iterations at which the moving average is performed as it tolerates larger stochastic oscillations. 

\subsection{Safeguarding}
To increase the robustness of our approach and ensure that the acceleration does not harm the training, we introduce a safeguarding checkpoint so that the accelerated step replaces the original one only if it further reduces the residual of the fixed-point iteration. 

The pseudo-code of the resulting AADL algorithm, combining moving average and safeguarding, is described as Algorithm \ref{AAR_algorithm}.

\begin{algorithm}[t]
\caption{Safeguarded AADL with moving average}
\label{AAR_algorithm}
\textbf{Input}: $\mathbf{w}^0$, $\boldsymbol{\xi}$\\
\textbf{Parameter}: $\beta$, $p$, $q$, $m$, $t$, $\epsilon$, $tol$\\
\textbf{Output}: $\mathbf{w}_{opt}$
\begin{algorithmic}[1]
\STATE{$\mathbf{r}^0 = G(\mathbf{w}^0, \boldsymbol{\xi}^0)-\mathbf{w}^0$} 
\STATE{$\mathbf{w}^1=\mathbf{w}^0 + \mathbf{r}^0$}
\STATE{$k=1$}
\WHILE{$\displaystyle \ell(\mathbf{w}^{k}, \boldsymbol{\xi}^k) >tol$}
 \STATE{$h=\min\{k,m\}$}
 \IF{$\|\sqrt{S^(\mathbf{w}^k)}\|_{\infty} > \varepsilon \cdot \|\mathbf{r}^k\|_{\infty}$}
\STATE{$\tilde{\mathbf{w}}^{k} = \frac{1}{t}\sum_{i=0}^{t-1}  \mathbf{w}^{k-i}$}\\
$\mathbf{w}^k = \tilde{\mathbf{w}}^{k}$
 \ENDIF
 \STATE{$\mathbf{r}^k = G(\mathbf{w}^k, \boldsymbol{\xi}^k)-\mathbf{w}^k$}
 \IF{$k\pmod q\ne 0$}
 \STATE{$W_k = [\mathbf{w}^{k-h+1} - \mathbf{w}^{k-h}, \ldots, \mathbf{w}^k-\mathbf{w}^{k-1} ] $}
 \STATE{$R_k = [\mathbf{r}^{k-h+1} - \mathbf{r}^{k-h}, \ldots, \mathbf{r}^k-\mathbf{r}^{k-1} ]$}
 \ENDIF
 \IF{$k\pmod p\ne 0$}
 \STATE{Set $\mathbf{w}^{k+1}=\mathbf{w}^k + \mathbf{r}^k$}
 \ELSE
 \STATE{Determine $\mathbf{g}^k = [g^k_1, \ldots, g^k_{
 \ell}]^T$ such that 
 $\displaystyle
 \mathbf{g}^k=\underset{\mathbf{g}\in \mathbb{R}^{\ell}}{\operatorname{argmin}}\lVert \mathbf{r}^k-R_k\mathbf{g}\rVert_2
 $}
 \STATE{Set $\mathbf{x}_{AA}^{k+1}=\big [ (1-\beta) I - \beta (X_k + R_k)(R_k^T R_k)^{-1} R_k^T \big ]  \mathbf{r}^k$}\\
 \STATE{Compute $\mathbf{r}^{k+1} = G(\mathbf{w}^{k+1}, \boldsymbol{\xi}^{k+1})-\mathbf{w}^{k+1}$}\\
 \STATE{Safeguarding:}
 \IF{$\lVert \mathbf{r}^{k+1} \rVert < \lVert \mathbf{r}^{k} \rVert$ }
 \STATE Set $\mathbf{w}^{k+1} = \mathbf{w}_{AA}^{k+1}$
 \ELSE
  \STATE{Set $\mathbf{w}^{k+1}=\mathbf{w}^k + \mathbf{r}^k$}
 \ENDIF
 \ENDIF
 \STATE{$k=k+1$}
\ENDWHILE
 \RETURN{$\mathbf{w}^{k+1}$ }
\end{algorithmic}
\end{algorithm}

\subsection{Computational cost of AADL}
Our implementation of AADL solves the least-squares problem in Equation \eqref{least_squares} by computing the QR factorization - see \cite{golub13}, Section 5.2 - of the $n\times m$ matrix $R_k$. 
The main components used in Algorithm \ref{AAR_algorithm} to perform AA with their computational costs expressed with the big-$O$ notation are:
\begin{itemize}
    \item stochastic first-order update with batch size $B$: $O(B n)$
    \item QR factorization to perform AA: $O(nm^2)$
    \item moving average: $O(nm)$
\end{itemize}
The most expensive step is AA, and its cost can be mitigated by increasing $p$. 
Denoting with $I$ the total number of stochastic updates the train the DL model, the computational complexity of the entire training is
\begin{equation}
    O\bigg( IBn + \bigg \lfloor \frac{I}{p}\bigg \rfloor nm^2\bigg),
\end{equation}
where $\lfloor \cdot \rfloor$ represents the highest integer smaller than $\frac{I}{p}$.
One way to restore the computational complexity of the first-order optimizer is to set $p = \frac{1}{m^2}$, but this may also result into relaxing the frequency to perform AA too much and the training may not benefit from the acceleration. 

\section{Numerical results}

We present numerical results that compare the performance of DL models trained with (i) unaccelerated state-of-the-art stochastic optimizers and with (ii) AADL to accelerate the training, highlighting the additional benefit provided by the moving average. 
Although empirical results have shown that an accurate hyperparameter tuning (e.g., learning rate) can improve the performance of a stochastic first-order optimizer, the hyperparameter configuration selected typically varies depending on the dataset \cite{choi2020on} and the type of hyperparameter optimization algorithm. Since this analysis goes beyond the scope of this work, we fix the values for the hyperparameters following general guidelines provided in the literature. Except for the \texttt{ImageNet1k}, the results are averaged over $20$ runs, and the random seed is fixed to an integer between $0$ and $19$ for each run. Since the \texttt{ImageNet1k} test case requires a much larger amount of computational resources and time, the results for this test case are shown by performing one single run. 

\subsection{Hardware description}
The numerical experiments are performed using Summit, a supercomputer at the Oak Ridge Leadership Computing Facility (OLCF) at Oak Ridge National Laboratory (ORNL). 
Summit has a hybrid architecture, and each node contains two IBM POWER9 CPUs and six NVIDIA Volta GPUs all connected together with NVIDIA’s high-speed NVLink. Each node has over half a terabyte of coherent memory (high bandwidth memory + DDR4) addressable by all CPUs and GPUs plus 1.6 TB of non-volatile memory (NVMe) storage that can be used as a burst buffer or as extended memory. To provide a high rate of communication and I/O throughput, the nodes are connected in a non-blocking fat-tree, using a dual-rail Mellanox EDR InfiniBand interconnect.

\subsection{Software description}
Our implementation uses \texttt{Python3.7} with the \texttt{PyTorch v1.3.1} package \cite{paszke_pytorch_2019} for auto-differentiation, and \texttt{torch.nn.parallel.DistributedDataParallel} is used to parallelize the training in a multi-process framework.
Our implementation of AADL is open-source and available at \cite{AADL}.

\subsection{Accelerated training of neural networks}

\subsubsection*{Graduate admissions dataset}
We present numerical results obtained by training an MLP on the open-source \texttt{graduate admissions} dataset \cite{graduate} that describe the probability of 500 students to be admitted to the Masters Program based on 9 parameters that measure their academic performance. The architecture of the MLP has 3 hidden layers, 64 neurons per hidden layer, and ReLU is used as activation function at each hidden layer. We use Adam \cite{adam} with an initial learning rate equal to $0.02$ that is reduced to $4\mathrm{e}-3$ after 1,000 epochs. 
The batch size is 40 data samples, the frequency to perform AA is $p=1$, the storage frequency is $q=1$, the history window is $m=10$, and the relaxation parameter is $\beta=0.1$. The adaptive criterion for the moving average uses a threshold $\varepsilon=0.1$.
The loss function minimized is the mean squared error (MSE).
The splitting of the dataset is 80\% for training and 20\% for validation. 
Figure \ref{UCI_graduate_admission} compares the validation MSE for Adam, Adam combined with AADL and Adam combined both with AADL and moving average. Adam combined with AADL performs similarly to the unaccelerated Adam, whereas the additional use of the moving average improves the training by increasing the final accuracy by an order of magnitude.  

\begin{figure}[ht]
\centering
\includegraphics[width=0.5\textwidth]{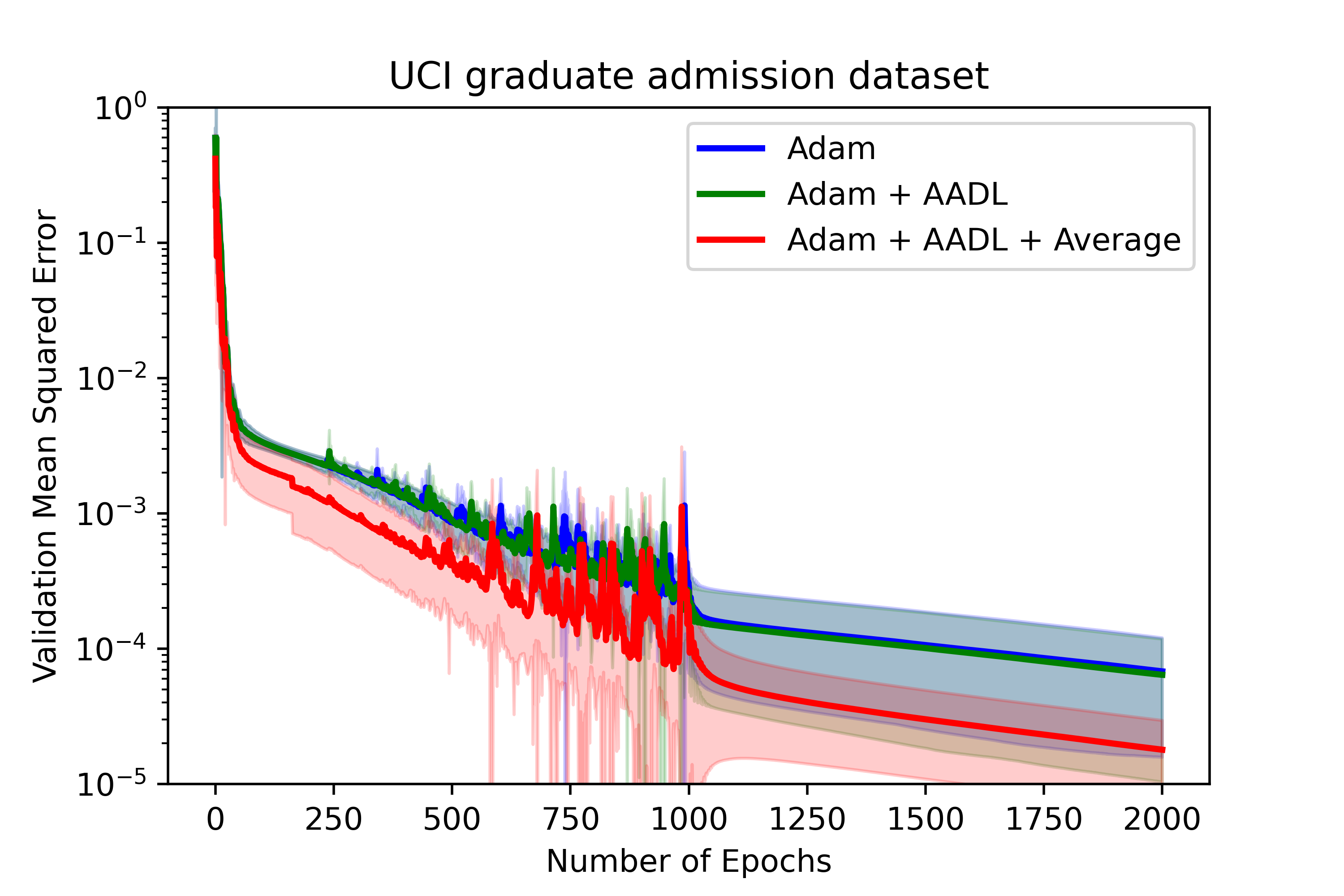}\caption{Validation MSE of an MLP trained on the \texttt{graduate admissions} dataset using the optimizer Adam (blue line), Adam combined with AADL (green line), and Adam combined with AADL and moving average (red line).
The solid line is the average over 20 runs and the shaded bands represent the 95\% confidence interval.}\hfill
\label{UCI_graduate_admission}
\end{figure}%

\subsubsection{$\mathbf{2d}$ and $\mathbf{100d}$ Burgers' equations}
The Burgers' equation defined as 
\begin{equation}
\begin{split}
    \mathcal{L}(u(\mathbf{x},t)) & =  u_t(\mathbf{x},t) + u(\mathbf{x},t) \bigg [ \sum_{a=1}^{d-1} u_{x_a}(\mathbf{x},t)\bigg ] \\ & - \nu \Delta u(\mathbf{x},t) = f(\mathbf{x},t)
\end{split}
\end{equation}
models the propagation of a nonlinear wave $u(\mathbf{x},t)$. The function $u(\mathbf{x},t)$ is the solution to the PDE, $u_t$ is the partial derivative of $u$ in time, $u_{x_a}$ is the partial derivative of $u$ in the $a$th dimension in space, and $\Delta u$ is the Laplacian operator. 
The integer $d$ denotes the dimensionality of the PDE which is either $d=2$ or $d=100$ for the test cases considered in this work, where one dimension is used for time and the remaining $d-1$ dimensions are used for space.

The PINN model is an MLP trained on source data to minimize the MSE, which is split in two parts:
\begin{equation}
    MSE = MSE_{u} + MSE_{f},
\end{equation}
where 
\begin{equation}
    MSE_{u} = \frac{1}{N_{u}} \sum_{i=1}^{N_u} \lvert u_{NN}(\mathbf{x}^i_u, t^i_u) - u_i \rvert^2
\end{equation}
and
\begin{equation}
    MSE_{f} = \frac{1}{N_{f}} \sum_{i=1}^{N_f} \lvert f(\mathbf{x}^i_f, t^i_f) - \mathcal{L}(u_{NN}(\mathbf{x}^i, t^i))  \rvert^2.
\end{equation}
Here $\{\mathbf{x}^i_u, t^i_u, u_i\}_{i=1}^{N_u}$ denote the initial and boundary training data on $u(\mathbf{x},t)$ and $\displaystyle\{\mathbf{x}^i_f, t^i_f\}_{i=1}^{N_f}$ specify the collocation points where the differential operator $\mathcal{L}$ is evaluated on the numerical solution $u_{NN}(\mathbf{x}^i_f, t^i_f)$ inside the domain at different points in time and space $\{\mathbf{x}^i_f, t^i_f\}_{i=1}^{N_f}$.
$MSE_f$ is the residual of the PDE inside the domain, and $MSE_u$ is the discrepancy between the numerical solution computed with PINN and the initial condition at time $t=0$, as well as the data on the boundary for any time $t$. We use $MSE_u$ to quantify the mismatch between the values of the approximate solution computed with the PINN model and the initial condition and boundary condition at different time steps $t_i$. We use $MSE_f$ to quantify the mismatch between the values of the left hand side $\mathcal{L}(u_{NN}(\mathbf{x}^i_f, t^i_f))$ and the right hand side $f(\mathbf{x}^i_f,t^i_f)$. The terms $MSE_u$ and $MSE_f$ are then combined to define the loss function that is minimized during the training of the PINN model. 

We first solve the $2d$ Burgers' equation, where the spatial domain is $[-1,+1]$, and the time interval is $[0 ,1]$. We denote the one-dimensional space variable with $x$. 
 The forcing term is set to $f(x,t)=0$ for any point $x$ in the spatial domain $(-1,+1)$ at any time $t$, and the viscosity parameter $\nu$ is set to $0.01/ \pi$. Homogeneous Dirichlet boundary conditions are imposed on the boundary of the spatial domain to force the solution to zero at any time $t$, and the initial condition is set to $u(x,0) = - \text{sin} (\pi x)$ for any $x$ in $(-1,+1)$. 
Each data batch samples 400 points $\{x^i_u, t^i_u, u_i\}_{i=1}^{N_u}$ to evaluate the initial condition at different $x^i_u$ inside the domain $(-1, +1)$ with $t_i = 0$, and the homogeneous Dirichlet boundary conditions at $x_i = -1$ and $x_i=+1$ at different values in time $t_i>0$. An additional set of 4,000 points $\displaystyle\{x^i_f, t^i_f\}_{i=1}^{N_f}$ are randomly sampled for different $x^i_f$'s inside the spatial domain $(-1,+1)$ and for different time steps $t^i_f > 0$, and the left hand side $\mathcal{L}(u_{NN}(x^i_f, t^i_f))$ of the Burgers' equation is evaluated at these points. 
The splitting of the data is 80\% for training and 20\% for validation. 
The MLP has two neurons in the input layer because 2 is the dimensionality of the PDE solved, followed by 8 hidden layers with 20 neurons each, and an output layer with 1 neuron since the solution $u$ is a scalar. We use ReLU as activation function for each hidden layer.
Adam is used as optimizer with an initial learning rate equal to $0.01$ that is reduced to by a factor of $0.2$ every 500 iterations. 
AADL is performed with a frequency $p=10$, a storage frequency is $q=5$, a history window is $m=5$, and the relaxation parameter is $\beta=1.0$. The adaptive criterion for the moving average uses a threshold $\varepsilon=0.1$.

Figure \ref{fig:lossburgers2D} shows the validation MSE when the training of the PINN model is performed with the unaccelerated Adam, Adam accelerated with AADL, and Adam combined both with AADL and moving average. AADL alone improves the performance by reducing the validation MSE. When the moving average is used, we notice intermediate oscillations that temporarily increase the validation MSE, which occur when a new data batch is selected and causes the moving average to mix gradient updates associated with the old and the new data batch. However, the numerical scheme promptly recovers from these temporary oscillations and improves the final accuracy.

\begin{figure}[ht]
 	\centering
 		\includegraphics[width=0.5\textwidth]{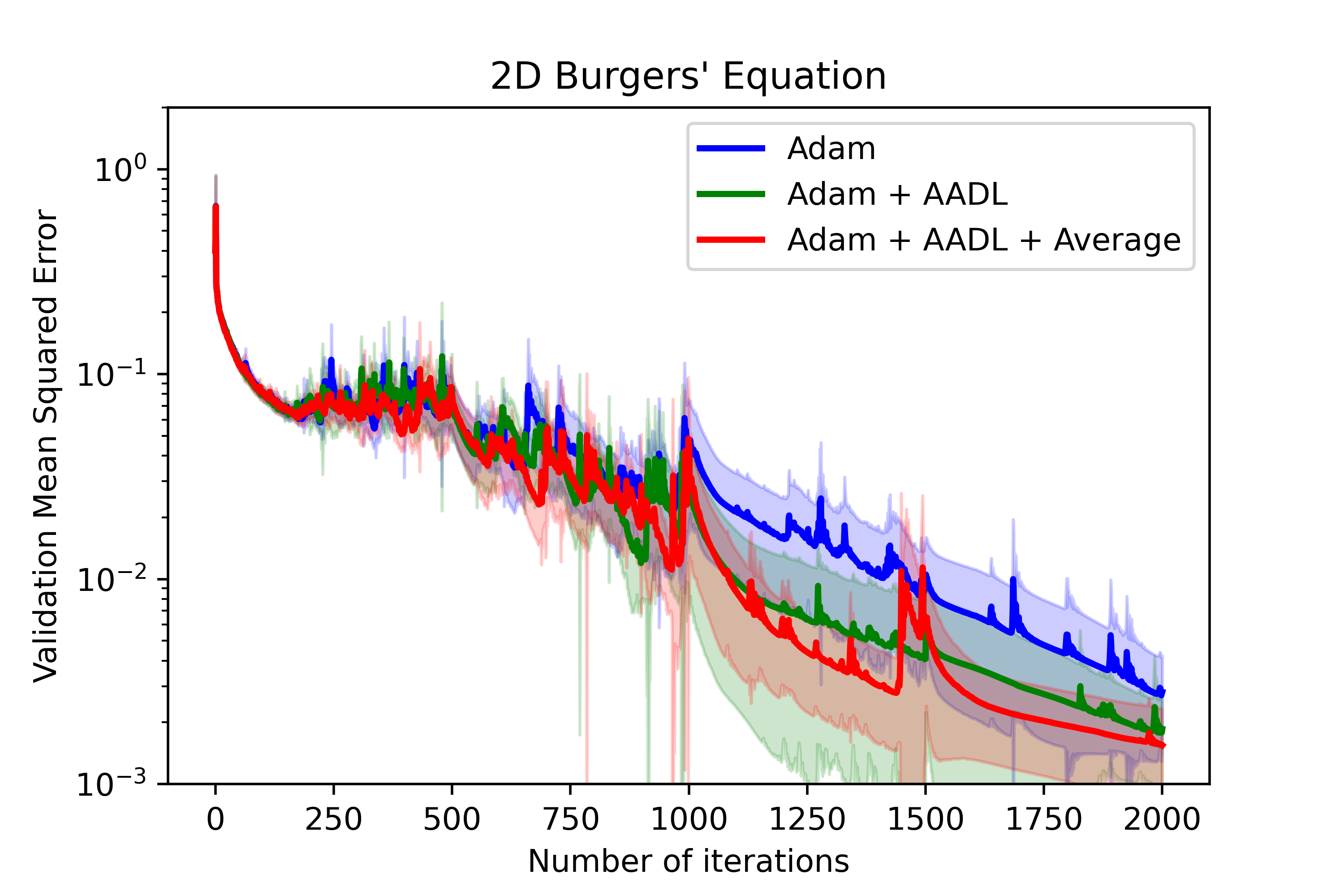}
 	\caption{Validation MSE for training the PINN model on source data for the $2d$ Burgers' equation using the optimizer Adam (blue line), Adam combined with AADL (green line), and Adam combined with AADL and moving average (red line). The solid lines are the average over 20 runs and the shaded bands represent the 95\% confidence interval.}
\label{fig:lossburgers2D}
 \end{figure}

We now use a PINN model to solve the $100d$ Burgers' equation, where the spatial domain is the hypercube $[-1,+1]^{99}$, the time interval is [0,1], and homogeneous Dirichlet boundary conditions are imposed.
We impose the exact solution to be
\begin{equation}
    u_{\text{exact}}(\mathbf{x},t) = e^{-t} (x_1 -1)(x_1 +1) \dots (x_n -1)(x_n + 1),
\end{equation}
and we construct the forcing term $f(\mathbf{x},t)$ that would produce this solution by evaluating the differential operator $\mathcal{L}$ on $u_{\text{exact}}(\mathbf{x},t)$. The initial condition is set to $u_{\text{exact}}(\mathbf{x},0)$. 
The data batch contains 400 points $\{\mathbf{x}^i_u, t^i_u, u_i\}_{i=1}^{N_u}$ that evaluate the initial condition at different $\mathbf{x}^i_u$ inside the domain $(-1, +1)^{99}$ with $t_i = 0$, and the homogeneous Dirichlet boundary conditions for different time values $t_i>0$. An additional set of 4,000 points $\displaystyle\{\mathbf{x}^i_f, t^i_f\}_{i=1}^{N_f}$ are randomly sampled for different values of $\mathbf{x}^i_f$ inside the spatial domain $(-1,+1)^{99}$ and for different time steps, and the left hand side $\mathcal{L}(u_{NN}(\mathbf{x}^i_f, t^i_f))$ of the Burgers' equation is evaluated at these points.
The 80\% of the data is used for training, and the remaining 20\% is used for validation. 
Adam is used as optimizer with an initial learning rate equal to $0.05$ that is reduced to by a factor of $0.5$ every 1,000 iterations. 
The MLP has an input layer with 100 neurons because 100 is the dimensionality of the PDE solved, followed by 4 hidden layers with 50 neurons each, and an output layer with 1 neuron since the solution $u$ is a scalar. AADL is performed at a frequency $p=1$, a storage frequency $q=1$, a history window $m=20$, and a relaxation parameter $\beta=1.0$. The adaptive criterion for the moving average uses a threshold $\varepsilon=0.1$.
Figure \ref{fig:lossburgers100D} shows the validation MSE to train the PINN model using the unaccelerated Adam, Adam accelerated with AADL, and Adam combined with AADL and the moving average. We notice that AADL alone does not improve the training, whereas the moving average efficiently reduces the oscillations of the training that occur when a new data batch is sampled, showing the stabilizing effect of the moving average over the training. 

\begin{figure}[ht]
 	\centering
 		\includegraphics[width=0.5\textwidth]{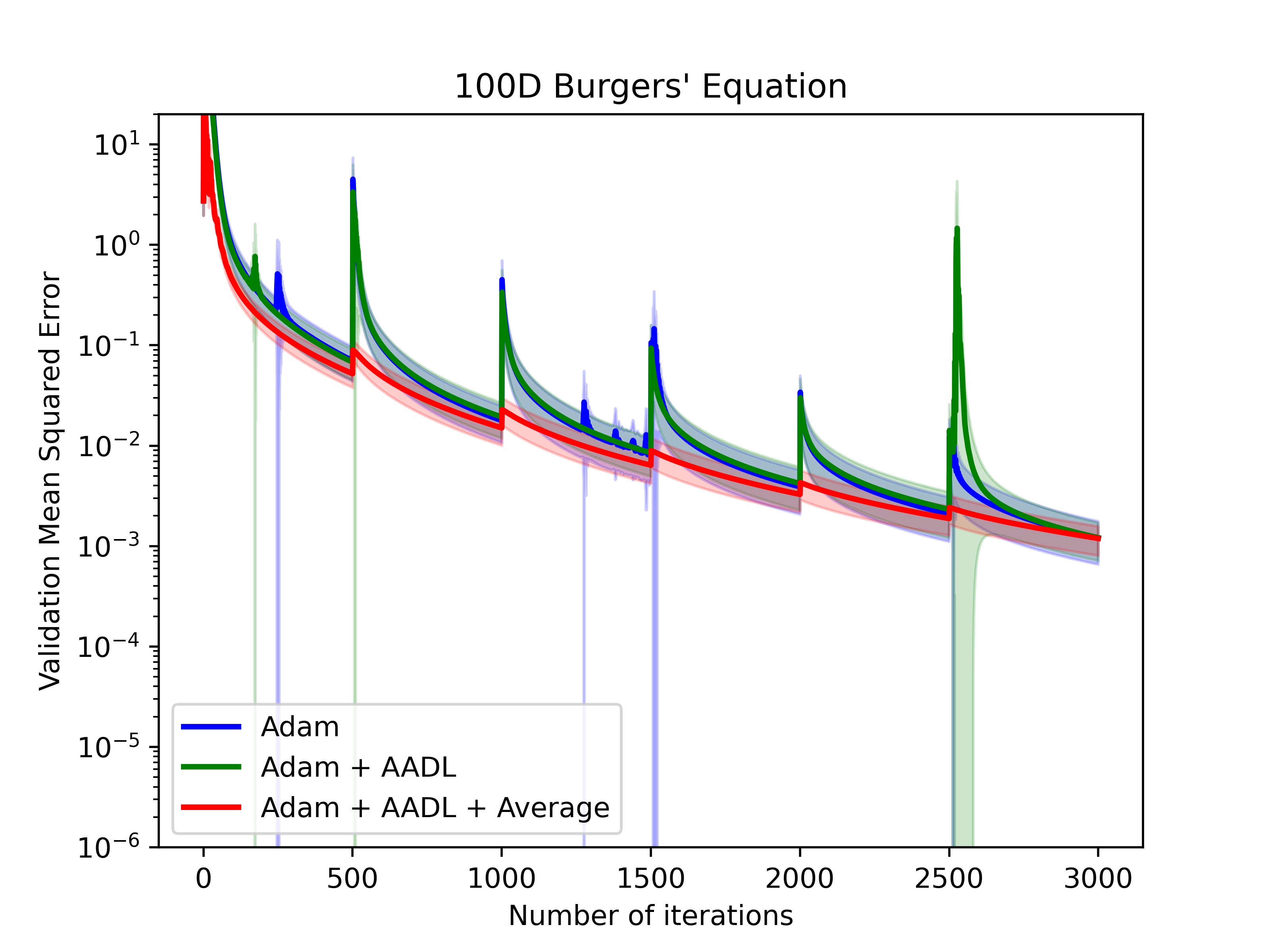}
 	\caption{Validation MSE for training the PINN model trained on source data for the $100d$ Burgers' equation using the optimizer Adam (blue line), Adam combined with AADL (green line), and Adam combined with AADL and moving average (red line). Frequency to perform AA set to $p=1$ and history window size set to $m=20$. The solid lines are the average over 20 runs and the shaded bands represent the 95\% confidence interval.}
\label{fig:lossburgers100D}
 \end{figure}

\subsubsection*{Image classification on ImageNet1k}
We train the ResNet50 model on the \texttt{ImageNet1k} dataset \cite{imagenet}, which consists of a training set of 1.2 million colored images and a test set of 50,000 colored images. Each image is labeled with one of 1,000 categories. The resolution of the images varies across the samples, with an average pixel size equal to 482 x 418. We use the binary cross-entropy as loss function to minimize during the training. 
We use Nesterov as stochastic first-order optimizer for training, using a momentum factor equal to 0.9 and 90 epochs for training. 
We perform data parallelism to distribute the training across multiple processes, and each process is linked to an NVIDIA V100 GPU. 
The data batch size is locally set to 64 for each process, and the total batch size is obtained multiplying the local batch size by the number of processes. We perform the parallel training at two different scale using 384 and 1,536 processes. The global batch size is thus equal to 24,576 for 384 processes and 49,152 for 1,536 processes.
We set the initial learning rate to 0.05 for the first 5 training epochs to warm-up the training, then the learning rate is set to a large initial value of 0.98 for 384 processes and 1.96 for 1,536 processes to compensate the averaging by across multiple processes. A learning rate scheduler decreases the value of the learning rate by a factor of 10 at epoch 30, 60, and 80. 
The frequency at which AA is performed is $p=5$, the frequency of storage is $q=5$, the history depth is $m=20$, and the relaxation parameter is $\beta=0.5$. 
The size of the ResNet50 model requires AA to be performed on the CPU host as the NVIDIA V100 GPUs do not have enough memory, introducing non-negligible communication overhead.
The numerical simulations compare the unaccelerated Nesterov against Nesterov accelerated with AADL, and Nesterov combined both with AADL and moving average.  
The unaccelerated Nesterov is affected by the phenomenon called generalization gap \cite{generalization}. AADL increases the computational time by a 10x factor without helping the optimizer recover from the generalization gap, as shown by the validation accuracy in Figure \ref{fig:imagenet_1536_accuracy} when 1,536 GPUs are used. Instead,  AADL combined with the moving average partially recovers from the generalization gap, showing that the moving average helps the DL model escape from bad local minima where the DL model under-performs.

\begin{figure}[ht]
 		\includegraphics[width=0.48\textwidth]{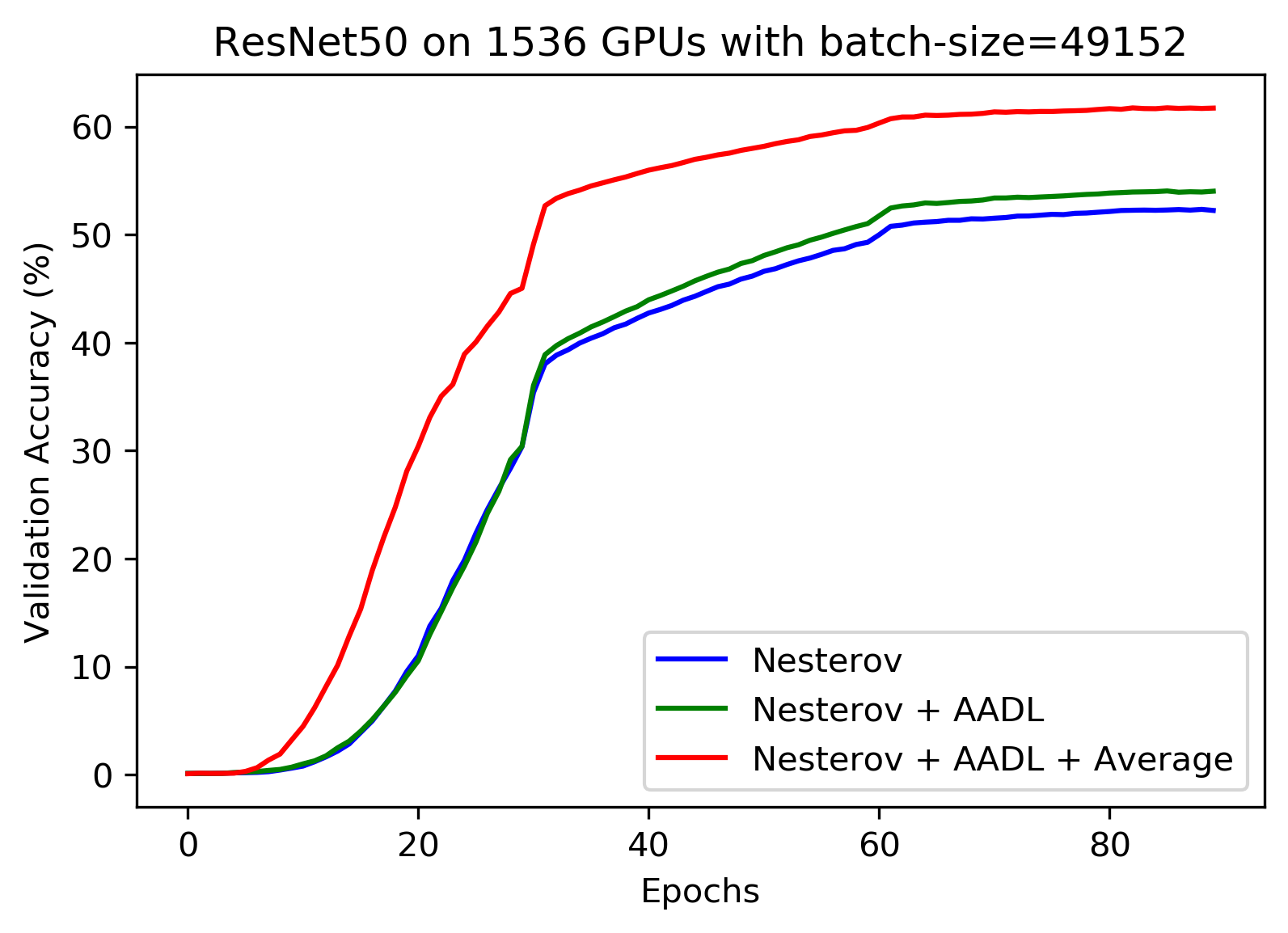}
 	\caption{Validation accuracy training the ResNet50 model on the \texttt{ImageNet1k} dataset with distributed data parallelism using 1,536 GPUs on 256 nodes of the OLCF supercomputer Summit.} 
 	 \label{fig:imagenet_1536_accuracy}
 \end{figure} 

\section{Conclusions and future work}
In this work we presented a moving average approach to stabilize the iterative training of DL models by damping the stochastic oscillations of the stochastic first-order optimizer, which allows to efficiently use AA for accelerates training. 
The numerical results show that the moving average helps AA stabilize and accelerate the convergence rate of the training in situation where AA alone is ineffective, as well as help AA recover from the generalization gap for large batch distributed training. 

Future work will focus on developing acceleration schemes that instead of being driven by the residual of the fixed point iteration are actually driven by quantities that directly measure the discrepancy between target data and predictions. Moreover, we will also extend the existing approach to situations where the architecture of the DL model is distributed across multiple GPUs. 

\section{Acknowledgments}
Massimiliano Lupo Pasini thanks Dr. Vladimir Protopopescu for his valuable feedback in the preparation of this manuscript.

This work is supported in part by the Office of Science of the US Department of Energy (DOE) and by the LDRD Program of Oak Ridge National Laboratory. This work used resources of the Oak Ridge Leadership Computing Facility (OLCF), which is a DOE Office of Science User Facility supported under Contract DE-AC05-00OR22725. 

\FloatBarrier

\bibliography{references}


\end{document}